\pdfoutput=1
\documentclass[11pt]{article}

\usepackage[final]{acl}

\usepackage{gb4e}
\noautomath
\usepackage{float}
\usepackage{paralist}
\usepackage{mathtools}
\usepackage[inline]{enumitem}
\usepackage{times}
\usepackage{latexsym}
\usepackage{amsmath}
\usepackage[capitalize]{cleveref}
\usepackage{amssymb}
\usepackage{multirow,multicol}
\usepackage{graphicx}

\usepackage{array}
\usepackage{bm}
\usepackage{soul, color}
\usepackage{booktabs}
\usepackage[colorinlistoftodos]{todonotes}
\usepackage{enumitem}
\usepackage{amsthm}
\usepackage{caption}
\usepackage{subcaption}
\usepackage{tipa}
\usepackage{stackengine}
\usepackage{rotating}
\usepackage{tabularx}
\usepackage[T1]{fontenc}
\usepackage[utf8]{inputenc}
\usepackage{longtable}
\usepackage{lipsum}
\usepackage{microtype}

\newcommand{\declarelogo}[0]{\includegraphics[height=.02\textwidth]{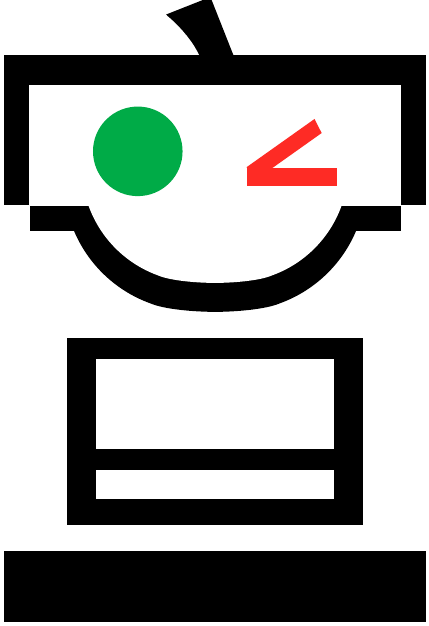}}
\newcommand{\umichlogo}[0]{\includegraphics[height=.012\textwidth]{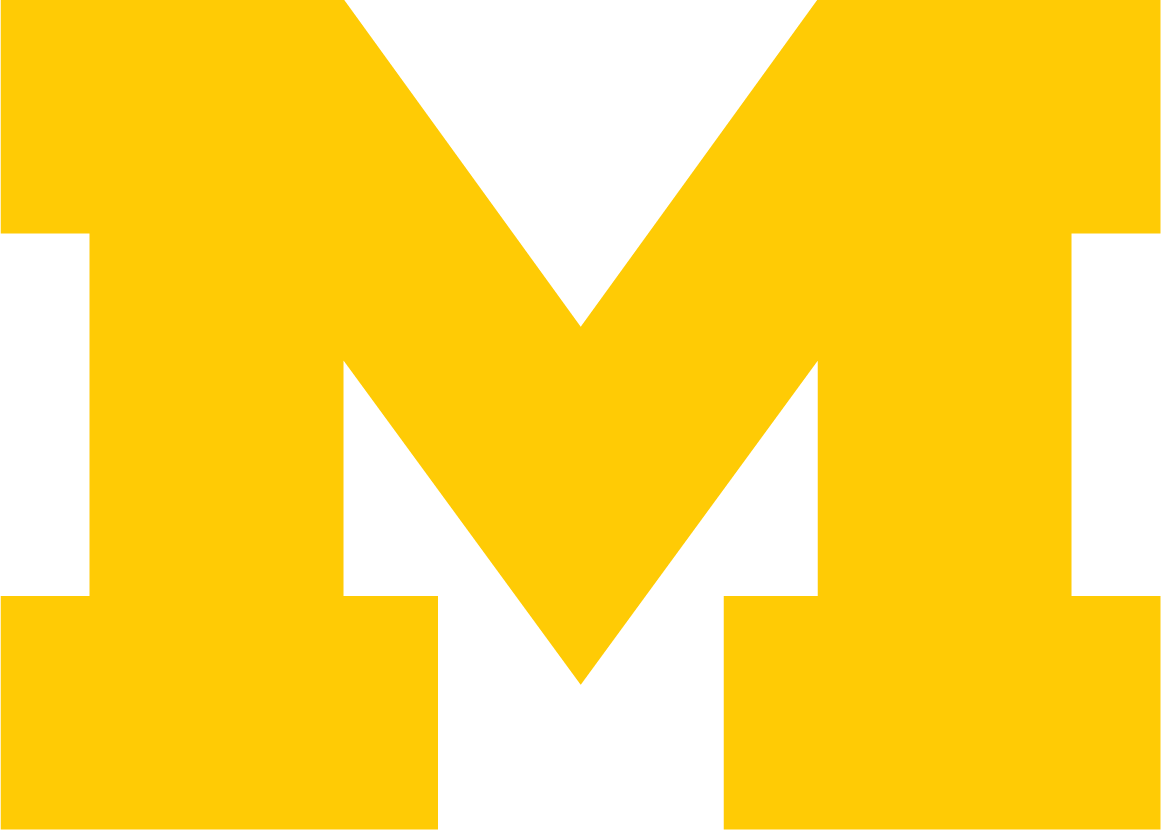}}

\newcommand\datatwofont[1]{{\usefont{T1}{cinzeldecorative}{m}{n}#1}}

\newcommand{\dataset}{{\datatwofont{CICERO$_{v2}$}}}
\newcommand{\model}{{DIALeCT}}
\newcommand{\PreserveBackslash}[1]{\let\temp=\\#1\let\\=\temp}
\newcolumntype{C}[1]{>{\PreserveBackslash\centering}p{#1}}
\newcolumntype{R}[1]{>{\PreserveBackslash\raggedleft}p{#1}}
\newcolumntype{L}[1]{>{\PreserveBackslash\raggedright}p{#1}}

\newcommand\code[1]{\texttt{#1}}

\crefformat{section}{\S#2#1#3} % see manual of cleveref, section 8.2.1
\crefformat{subsection}{\S#2#1#3}
\crefformat{subsubsection}{\S#2#1#3}
\definecolor{darkpastelgreen}{rgb}{0.01, 0.75, 0.24}
\definecolor{brilliantlavender}{rgb}{0.96, 0.73, 1.0}
\definecolor{Gray1}{rgb}{0.91,0.925, 0.937}
\definecolor{cambridgeblue}{rgb}{0.0, 0.8, 0.6}
\definecolor{Gray2}{rgb}{0.87, 0.886, 0.902}
\definecolor{Gray3}{rgb}{0.808, 0.831, 0.855}
\definecolor{Gray4}{rgb}{0.678,0.71, 0.741}
\definecolor{darkgreen}{rgb}{0.0, 0.5, 0.0}
\definecolor{almond}{rgb}{0.99, 0.87, 0.9}
\definecolor{ghostwhite}{rgb}{0.98, 0.81, 0.69}
\definecolor{Blue1}{rgb}{0.792, 0.941, 0.973}
\definecolor{Blue2}{rgb}{0.678, 0.91, 0.957}
\definecolor{Blue3}{rgb}{0.565, 0.878, 0.937}
\definecolor{Blue4}{rgb}{0.282, 0.749, 0.89}
\definecolor{Yellow1}{rgb}{1, 0.914, 0.306}
\definecolor{Yellow2}{rgb}{1, 0.886, 0.275}
\definecolor{Yellow3}{rgb}{1, 0.855, 0.239}
\definecolor{gold}{rgb}{0.85,.66,0}
\definecolor{aqua}{rgb}{0.80784314, 0.90196078, 0.35294118}
\definecolor{Green0}{rgb}{0.909, 0.992, 0.886}
\definecolor{Green1}{rgb}{0.843, 0.960, 0.839}
\definecolor{Green2}{rgb}{0.635, 0.854, 0.627}
\definecolor{amethyst}{rgb}{0.6, 0.4, 0.8}
\definecolor{persianpink}{rgb}{0.97, 0.5, 0.75}
\definecolor{unitednationsblue}{rgb}{0.67, 0.78, 0.98}

\definecolor{Amber}{rgb}{0.99, 0.76, 0.8}
\definecolor{bluebell}{rgb}{0.64, 0.64, 0.82}

\title{Pre-training Text-to-Text Transformers for Multiview Contextual Commonsense Inference in Dialogues}
\title{Multiview Contextual Commonsense Inference: A New Dataset and Task}

\author{First Author \\
  Affiliation / Address line 1 \\
  \texttt{email@domain} \\\And
  Second Author \\
  Affiliation / Address line 1 \\
  \texttt{email@domain} \\}

\author{
Siqi Shen$^{\bigstar\umichlogo}$ \hspace{2mm}
Deepanway Ghosal$^{\bigstar\declarelogo}$ \hspace{2mm}
Navonil Majumder$^{\declarelogo}$ \hspace{2mm} Henry Lim$^{\declarelogo}$ \hspace{2mm} \\
\textbf{Rada Mihalcea$^{\umichlogo}$ \hspace{2mm}
Soujanya Poria$^{\declarelogo}$}\\
  $^{\umichlogo}$ University of Michigan, USA\\
  $^{\declarelogo}$ DeCLaRe Lab, Singapore University of Technology and Design, Singapore\\
  \texttt{\{shensq,mihalcea\}@umich.edu}\\
  \texttt{deepanway\_ghosal@mymail.sutd.edu.sg}\\
  \texttt{\{navonil\_majumder@,henry\_lim@,sporia@\}sutd.edu.sg}\\ 
 \vspace{1mm}\code{\textbf{\dataset{} is available at: \url{https://declare-lab.github.io/CICERO}}}
  }

\begin{document}
\maketitle
\begin{abstract}
Multiview contextual commonsense inference is the task of determining commonsense explanations around the events in a dyadic dialogue, where multiview refers to the characteristic that there can be multiple plausible but independent inferences. Producing a coherent and non-trivial explanation requires awareness of the dialogue's structure and how an event is grounded in the context, yet there is a lack of high-quality resources dedicated to the task.  
In this work, we create \dataset{}, a dataset consisting of 8,351 instances from 2,379 dialogues, containing multiple human-written answers for each contextual commonsense inference question, representing a type of explanation on cause, subsequent event, motivation, and emotional reaction. We show that the inferences in \dataset{} are of higher semantic diversity than other contextual commonsense inference datasets. 
In addition, we propose a collection of pretraining objectives, including concept denoising and utterance sorting, to help adapt language models for the multiview contextual commonsense inference task. Evaluation results show the effectiveness of the pretraining stage, as there is a universal improvement in accuracy for all inference types. 
\end{abstract}

\section{Introduction}

%\blfootnote{$\bigstar$ Equal Contribution}
Perhaps unwittingly, commonsense is a key part of daily conversations. Rather than being explicit, interlocutors usually rely on shared context and commonsense knowledge to make sense of the inbound utterances and respond as succinctly as possible to maximize information flow~\cite{grice75logic}. The scope of this shared context, however, is quite often broad enough to span beyond the scope of the given conversation. Understanding various dimensions of such conversations for NLP systems is thus rather challenging without the aid of commonsense-based reasoning. Some of the useful dimensions, such as cause, subsequent events, and motivation behind some given utterance, can be extracted from the explicit context. Otherwise, the broader context that fits the explicit context must be imagined. Either way, commonsense knowledge must be employed with the context in mind to broaden the context if necessary and arrive at a fitting explanation. Inferring such explanations for various dimensions with the context and commonsense-based reasoning is called contextual commonsense inference. An accurate understanding of dialogues achieved through contextual commonsense inference can assist in meaningful indexing, filtering, and searching of the copious amount of conversational content available on the internet. Tasks like affect analysis and relation extraction in dialogues may also benefit from such explanations.

To this end, the CICERO dataset~\cite{ghosal-etal-2022-cicero} collects five dimensions of contextual commonsense inferences for utterances in dialogues. However, for each present dimension-utterance pair, only one human-annotated explanation is collected. The remaining explanations, if any, are picked using adversarial filtering~\cite{zellers2018swag} from a set of fine-tuned language model-generated explanations. These auto-generated explanations are both lexically and semantically very close to the human-annotated explanation. This contradicts the intuitive multiview nature of these explanations, where multiple disparate explanations for the same event may exist (see \cref{fig:showcase}). \dataset{} seeks to address this issue by collecting multiple distinct human-annotated explanations, leading to the enrichment of the downstream models for contextual commonsense inference task.

\looseness=-1 The availability of multiple correct answers brings the need for methods that can simultaneously select multiple correct answers from a mixture of correct and incorrect answers given a context. \citet{ghosal-etal-2022-cicero} shows that given a context, selecting two correct answers is harder than selecting just one. On CICERO, T5-Large attains an Exact Match (EM) score of 95\% on the single answer selection task but this score drops to 20\% on the multiple answer selection task. Models need to encode rich commonsense knowledge to solve this task due to its hardness. In this work, we attempt to encode commonsense knowledge to a large pre-trained language model T5-Large by continuing training it on a dialogue-level commonsense dataset CICERO~\cite{ghosal-etal-2022-cicero} using a set of commonsense-aware pre-training objectives.
Large pre-trained language models, such as GPT-2~\cite{Radford2019LanguageMA} and T5~\cite{Raffel2020ExploringTL}, seem attractive frameworks to solve contextual commonsense inference task. Through fine-tuning, these models have become state of the art in several natural language understanding tasks, such as SuperGLUE~\cite{Wang2019SuperGLUEAS}. Additionally, being trained on several hundreds of GB of text may have endowed these models with much commonsense knowledge~\cite{Petroni2019LanguageMA}. 

However, the fine-tuning approach may not suffice for tasks with limited training samples. Nonetheless, previous work ~\cite{Gururangan2020DontSP,Zhou2021PretrainingTT} has shown that, prior to fine-tuning, pre-training with objectives catered to the target tasks may improve performance on such tasks. Following this intuition, we propose a set of self-supervised pre-training objectives
% , which are collected without any further human annotation,
to adapt the language models for the contextual commonsense inference task, specifically addressing the task of multi-choice answer selection.  

Thus, our contribution in this paper is twofold: \emph{i)} we curate \dataset{}, containing multiple distinct contextual commonsense inferences per dimension, and
\emph{ii)} we propose a set of pre-training objectives for contextual commonsense inference that improves over the vanilla fine-tuning by about 1.9\% for the multi-choice answer selection task, defined on both CICERO and \dataset{} datasets.

\section{Primer on CICERO}
\looseness=-1 The dialogues in CICERO~\cite{ghosal-etal-2022-cicero} are sourced from three different datasets: DailyDialog~\cite{li2017dailydialog}, MuTual~\cite{mutual}, and DREAM~\cite{sun2019dream}. All  dialogues are dyadic and their inherent nature is particularly conducive to qualitatively rich utterance-level inferences. These annotated inferences  are categorized into five dimensions: cause, subsequent event, prerequisite, motivation, and emotional reaction. The tasks proposed on these inferences require contextual understanding, multi-utterance reasoning, and commonsense knowledge.

In addition to introducing CICERO, \citet{ghosal-etal-2022-cicero} also defines a multi-choice answer selection task (MCQ), where the original annotation is considered as the primary correct answer. The candidates for the remaining correct and incorrect answers are generated using  fine-tuned T5 models~\cite{raffel2019exploring}. Adversarial filtering~\cite{zellers2018swag} is applied to these candidates to identify the hard-to-distinguish answers, which are  manually labeled as correct or incorrect.

\begin{figure*}[ht]
    \centering
    \includegraphics[width=\textwidth]{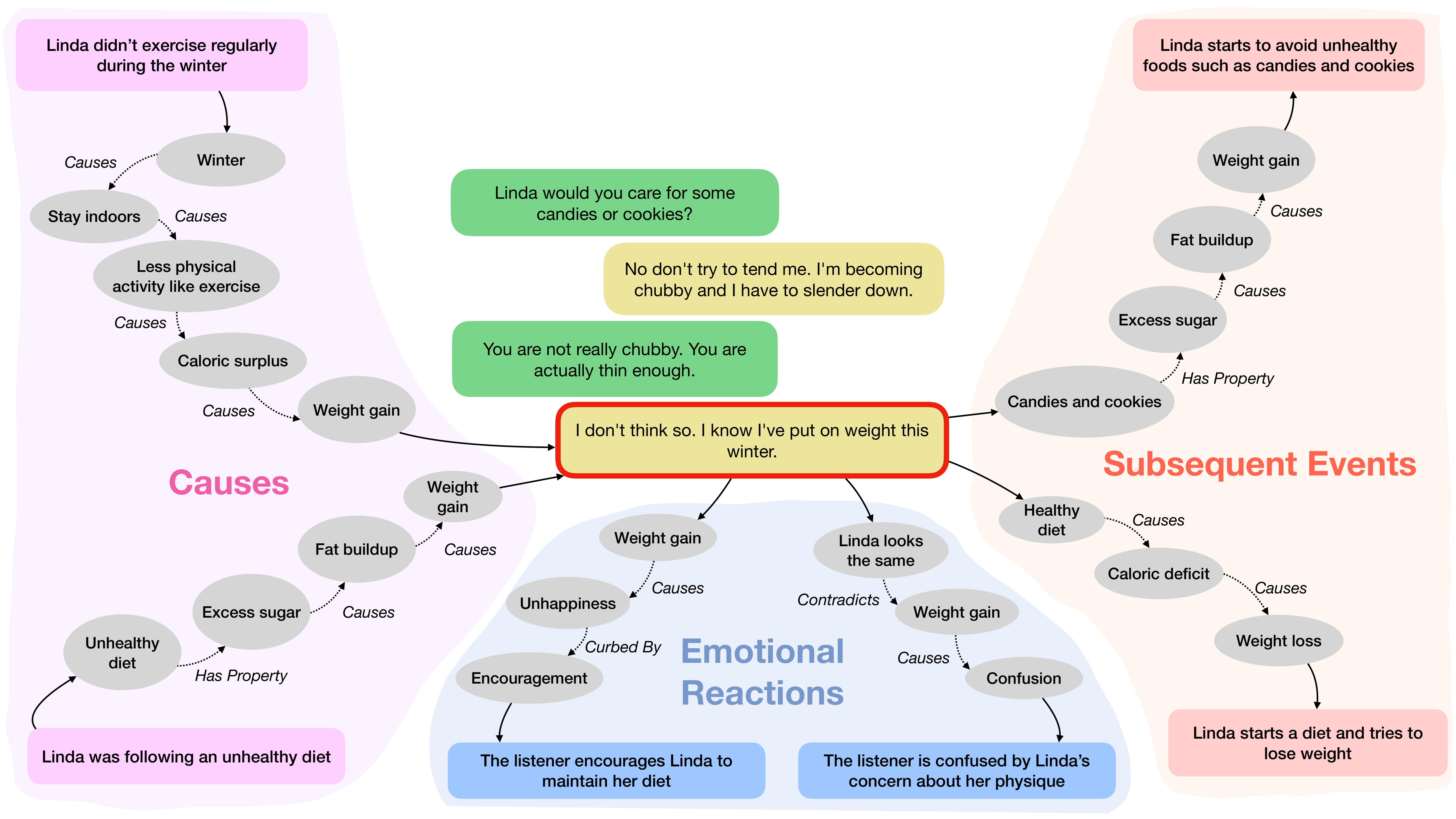}
    \caption{\footnotesize Demonstration of multiple possible contextual explanations through multiple commonsense-based mechanisms.}
    \label{fig:showcase}
\end{figure*}

\paragraph{Drawbacks of CICERO.} The automatically-generated and labeled-as-correct answers are the only sources of secondary correct answers in the CICERO dataset. In total, close to 15\% of the instances contain multiple correct answers (inferences). We empirically analyzed these instances and found that the adversarial filtering algorithm favors the selection of alternate answers that are lexically close to the primary correct answer. As such, both correct and incorrect answers bear a relatively high degree of token-level and semantic similarity with each other as indicated in \cref{tab:sim} in terms of BLEU, ROUGE-L, CIDER and semantic-similarity metrics. This belies the multiview nature of commonsense-based explanations, where multiple either independent or related explanations of the same event may exist. This is demonstrated in \cref{fig:showcase} where the target utterance ``\emph{I don't think so. I know I've put on weight this winter.}'' can be a consequence of multiple possible events. Particularly, the event of weight gain can be caused by lack of physical activity and exercise or unhealthy diet or perhaps both. There are myriad of other possible factors that may contribute to the weight gain, such as disease, but those multitudes of possibilities or views are not captured in CICERO. 

\section{CICEROv2}
\label{sec:dataset}

To address the drawbacks highlighted earlier, we introduce 
\dataset{},  to improve the generalization ability of the models trained on this data. 
\dataset{} contains commonsense inferences from target utterances of dyadic dialogues sampled from CICERO. A human annotator is given a \textit{dialogue} with a \textit{target} utterance and asked a \textit{question} about the \textit{target} utterance. The annotator writes multiple distinct correct answers and two or more incorrect answers for the question. 

% Nearly 70\% of the instances in \dataset{} are extended from the annotations in CICERO.

We start by sampling (\textit{dialogue}, \textit{target}, \textit{question}) triplets from CICERO. For these instances, we show the original correct answer from CICERO to the annotators to avoid duplication. The annotators write at least one more correct and at least two incorrect answers that are semantically distinct from each other and the answer from CICERO. This original answer and the newly written answer(s) constitute the set of answers for these instances. 

% The other 30\% of the instances are newly sampled
We also sample new
(\textit{dialogue}, \textit{target}, \textit{question}) triplets, not present in CICERO. The annotators write at least two correct answers and two incorrect answers for these instances. 

The above strategy ensures that all instances in \dataset{} have at least two correct and two incorrect answers. 

\subsection{Annotation Instructions}

\paragraph{Guidelines for Writing Correct Answers.}

We instruct the annotators to write context-congruent correct answers that are grammatically sound and concise sentences. The answers may contain some important terms from the context and must be commonsense-based, factual, and plausible.

\paragraph{Guidelines for Writing Incorrect Answers.}
The incorrect answers are also grammatically correct and concise but must contradict some information in the dialogue. Incorrect answers should contain some important terms from the context and must be commonsense-based and factual. Annotators were instructed not to write incorrect answers that are clearly outlandish in the given context.

We also ask the annotators to write sufficiently diverse and distinct correct and incorrect answers. This diversity may stem from token-level differences, semantic differences, or various likely speculative scenarios around the given context. Human-written diverse incorrect answers is a major contribution in \dataset{}, which is absent CICERO. 
We discuss the diversity of answers in CICERO and \dataset{} in more detail in \cref{sec:diversity}.

We collect inferences across four different dimensions in \dataset{}: \textit{subsequent event}, \textit{cause}, \textit{motivation}, and \textit{emotional reaction} w.r.t the \textit{target}. \emph{Prerequisite} dimension from CICERO is skipped as the annotators found it difficult to distinguish from \emph{cause} during annotation training. The annotators are asked to write correct and incorrect answer(s) to the questions representing each of the four inference dimensions. We expand on the annotation instructions outlined by \citet{ghosal-etal-2022-cicero} for answer writing. Both correct and incorrect answers may describe either an \emph{overt} or a \emph{speculative} scenario, as illustrated in CICERO. An overt answer is explicitly or implicitly present in the dialogue context. However, when a dialogue does not explicitly or implicitly hold the answer to a \textit{question} about a particular \textit{target}, the answer is speculated within the dialogue context imagined and broadened using commonsense and world knowledge. 

\begin{figure}[t]
    \centering
    \includegraphics[width=\linewidth]{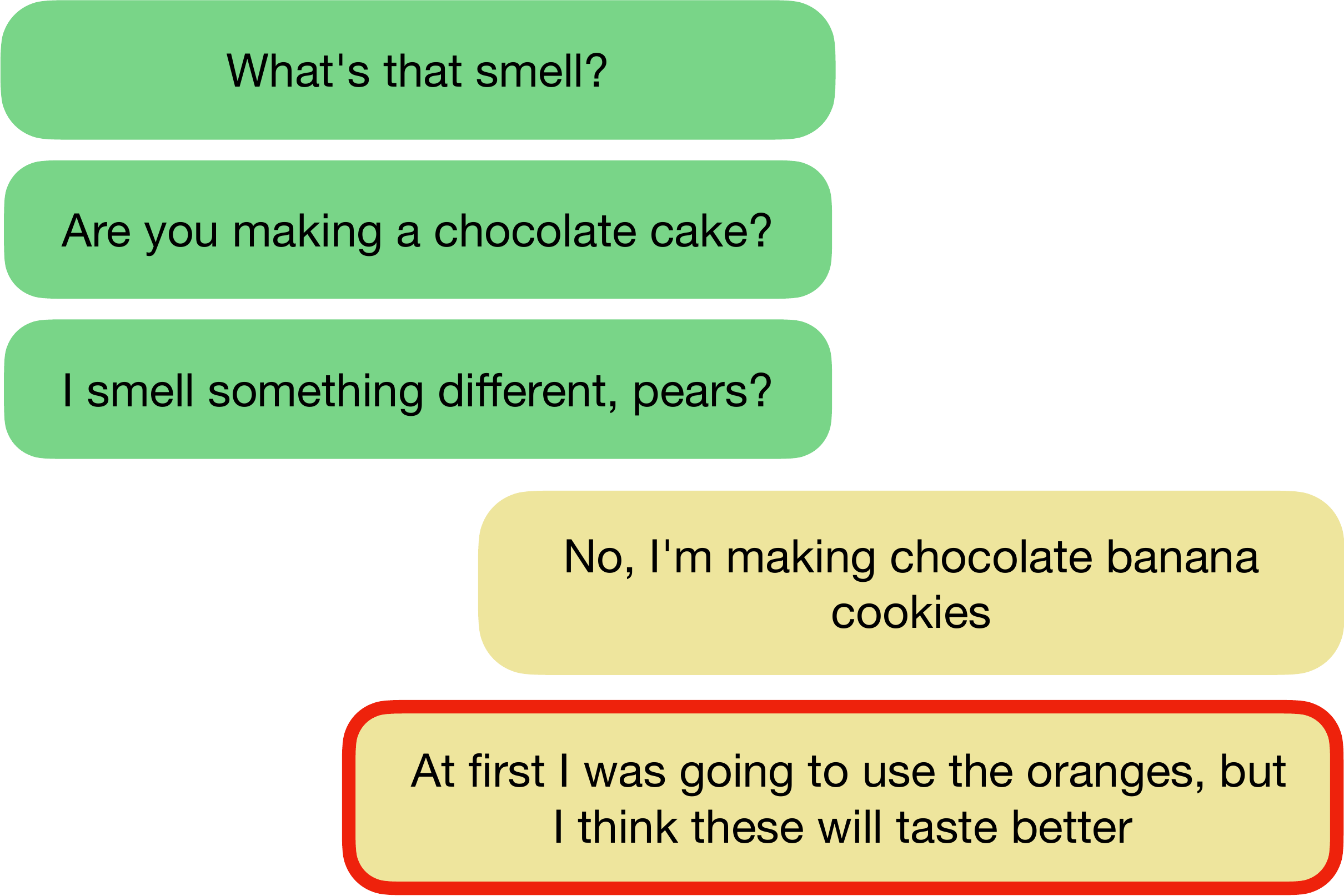}
    \caption{\footnotesize A (\textit{dialogue, target}) pair; the utterance with the red border is the \textit{target}.}
    \label{fig:dialogue-target-1}
\end{figure}

The following illustrates the \textit{questions} and possible correct and incorrect answer(s) for the (\textit{dialogue}, \textit{target}) pair shown in \cref{fig:dialogue-target-1}.
% We show only one incorrect answer for illustration here.

\paragraph{Q1.} \textbf{What subsequent event happens (overt) or could happen (speculative) following the \textit{Target}?}
The annotators write about the event that happens or could happen following the \textit{target}. They are also made aware that at times such subsequent events could be triggered by the \textit{target} itself.

\noindent{\bf CICERO Correct Answer:} The speaker made delicious banana cookies.
\noindent{\bf Incorrect Answers:} i) The speaker is making a chocolate cake. ii) The speaker was baking a cake.

\noindent{\bf \dataset{} Correct Answer:} The speaker threw the leftover oranges into the rubbish bin.
\noindent{\bf Incorrect Answers:} i) The listener requests to taste the orange cookies. ii) The listener started to make orange chocolate cookies.

\paragraph{Q2.} \textbf{What is the event that directly causes (overt) or could cause (speculative) \textit{Target}?} The annotators consider the events antecedent to the \textit{target} that cause or likely cause the \textit{target}.

\noindent{\bf CICERO Correct Answer:} The speaker was making banana cookies.
\noindent{\textbf{Incorrect Answers:}} i) The speaker is making a chocolate cake. ii) The speaker was baking a cake.

\noindent{\bf \dataset{} Correct Answers:} i) It is too difficult to process the orange pulp. ii) The orange smell doesn't match well with chocolate. 
\noindent{\bf Incorrect Answers} i) The orange smell matches much better with chocolate compared with banana. ii) The speaker loves the taste of orange and the texture of its pulp.

\paragraph{Q3.} \textbf{What is the emotion or basic human drive that motivates or could motivate \textit{Target}?}
We ask the annotators to consider the basic human drives and needs of the speaker of the \textit{target} utterances. The basic human drives include food, water, clothing, rest, safety, friends, relationships, enjoyment, etc. Do or may any of the human drives/states of mind/emotional feelings motivate the \textit{target}?

\noindent{\bf CICERO Answers:} \textit{Instance not present.}

\noindent{\bf \dataset{} Correct Answers:} i) The speaker wants the cookies to be delicious. ii) The oranges were not sweet enough for the cookies.
\noindent{\bf Incorrect Answers:} i) The speaker prefers spicy cookies. ii) The speaker wants to use the leftover pears before they go bad. 

\paragraph{Q4.} \textbf{What is the possible emotional reaction of the listener: A (or B)?}
What could be the possible emotional reaction of the listener to the \textit{target}? The annotators capture the appropriate emotion of the listener using the emotion terms listed in the Appendix in \cref{tab:emotions} using verbatim or related words (e.g., anxious, confused, interested). 

\noindent{\bf CICERO Correct Answer:} The listener is excited to eat the cookies.
\noindent{\bf Incorrect Answers} i) The listener is excited to eat the salad. ii) The listener is excited to eat the muffins instead.

\noindent{\bf \dataset{} Correct Answer:} The listener feels pity that they cannot have orange cookies.
\noindent{\bf Incorrect Answers} i) The listener is happy to taste orange cookies. ii) The listener is annoyed by the banana smell.

\subsection{Sampling of Dialogues and Targets}
From the (\textit{dialogue}, \textit{target}, \textit{question}) triplets in CICERO, the following criteria is used to subsample a set of triplets for annotation:

\vspace{-\topsep}
\begin{itemize}[leftmargin=*, wide, itemsep=0pt, labelwidth=!, labelindent=0pt]
\setlength\itemsep{-0.25em}
    \item The \textit{target} utterance must contain at least one non-stop verb word and more non-stop words than stop words.
    \item If the \emph{dialogue} is from DailyDialog, then the dialogue-act label of the \textit{target} utterance must either be directive or commissive~\cite{li2017dailydialog}. 
\end{itemize}
\vspace{-\topsep}

These sampled target utterances often describe some action or activity, which the annotators found easier to annotate across the four question types. Overall, 17\% of the correct answer annotations in \dataset{} also appear in CICERO. However, there is no overlap between the incorrect answers in the two datasets. Crucially, \dataset{} contains all manually annotated and semantically diverse set of commonsense-based correct and incorrect answers that capture distinct perspectives or views. We expand upon the diversity of the answers next.

% 70\% of the triplets in \dataset{} are directly imported from CICERO, for which the annotators write additional correct and incorrect answers. The remaining 30\% had their \emph{question} changed from CICERO such that the new triplets are not present in CICERO.

\subsection{Diversity of Answers}
\label{sec:diversity}
\begin{table}[t]
\small
\centering
\resizebox{0.75\linewidth}{!}{
%\scalebox{0.6}{
	\begin{tabular}{p{4.7cm}@{}|c@{~~}|c@{~~}}
	\toprule
	\textbf{Description} & \textbf{\dataset{}} & \textbf{CICERO}\\
	\midrule
	\bf \# Dialogues / \# Instances & & \\
    $\quad$ DailyDialog &  1,118 / 3,973 & 2,113 / 4,344  \\
    $\quad$ MuTual &  1,011 / 3,384  &  929 / 1,715  \\
    $\quad$ DREAM &  250 / 994 &  516 / 1,386 \\
    $\quad$ \bf Total &  2,379 / 8,351  & 3,558 / 7,445 \\
    \midrule
    \# \bf Dialogues with \# Instances & & \\
    $\quad$ $<$ 4 & 1,377 & 3,057 \\
    $\quad$ 4 $\leq * \leq$ 8 & 919 & 493 \\
    $\quad$ $>$ 8 & 83 & 8\\
    % \bf Avg. \# Inferences per Dialogue &  & --\\
    \midrule
    % \bf Instances with
    % \bf \# Total Answers & & \\
    % $\quad$ $=$ 4 & 4541 &  \\
    % $\quad$ $=$ 5 & 2456 &  \\
    % $\quad$ $>$ 5 & 52 &  \\
    \bf Avg. \# of Correct Answers & 2.40 & 2.49 \\
    \midrule
    \bf Instances with \# Correct Answers & & \\
    $\quad$ $=$ 2 & 5,066 & 4,985 \\
    $\quad$ $=$ 3 & 3,260 & 1,552 \\
    $\quad$ $>$ 3 & 25 &  908 \\
    \midrule
    \begin{tabular}{l}
    \bf Question Types in \\ \bf Train / Validation / Test \end{tabular} & &  \\
    $\quad$ Cause &  1,227 / 189 / 243 &  1,301 / 381 / 514  \\
    $\quad$ Subsequent Event & 2,196 / 618 / 793 & 1,193 / 568 / 759  \\
    $\quad$ Motivation & 1,457 / 330 / 480  &  455 / 163 / 194  \\
    $\quad$ Reaction & 561 / 116 / 141 & 234 / 105 / 116 \\
    $\quad$ Prerequisite & - & 1,010 / 201 / 251 \\
    % $\quad$ \bf Total & & \\
    \bottomrule
	\end{tabular}
	}
	\caption{\footnotesize Statistics of \dataset{}. We report the numbers only for the multiple correct answer subset in CICERO.}
	\label{tab:stat}
\end{table}

\begin{table}[h]
% \small
\centering
\resizebox{\linewidth}{!}{
\begin{tabular}{c|cccccc}
\toprule
\textbf{Data (x, y)} & \textbf{BLEU1} & \textbf{BLEU2} & \textbf{BLEU4} & \textbf{ROUGE-L} & \textbf{CIDER} & \textbf{Sem-Sim}\\
\midrule
% v1 & C & C & 0.4817 & 0.7323 & 0.2918 & 0.7974 \\
% v1 & I & I & 0.3442 & 0.6119 & 0.7434 & 0.7120 \\
% v2 & C & C & 0.0604 & 0.3636 & 0.0049 & 0.5964 \\
% v2 & I & I & 0.0777 & 0.3907 & 0.0016 & 0.5516 \\
v1 (C, C) & 0.7082 & 0.6340 & 0.4817 & 0.7323 & 0.2918 & 0.7974 \\
v1 (I, I) & 0.5966 & 0.5036 & 0.3442 & 0.6119 & 0.7434 & 0.7120 \\
v1 (C, I) & 0.6797 & 0.6028 & 0.4565 & 0.7016 & 0.1268 & 0.7355 \\
v2 (C, C) & 0.3265 & 0.1966 & 0.0501 & 0.3533 & 0.0028 & 0.5934 \\
v2 (I, I) & 0.3455 & 0.2164 & 0.0625 & 0.3738 & 0.0009 & 0.5425 \\
v2 (C, I) & 0.3367 & 0.2214 & 0.0685 & 0.3614 & 0.3421 & 0.5097 \\
\bottomrule
\end{tabular}
}
\caption{\footnotesize (x, y) indicates source-target pair. v1, v2, C, I indicate CICERO, \dataset{}, correct answer set, and incorrect answer set, respectively. We show the instance-level average similarity between pairs of (correct, correct), (incorrect, incorrect), and (correct, incorrect) answers in CICERO and \dataset{}.}
\vspace{-1em}
\label{tab:sim}
\end{table}
Answers in \dataset{} are significantly more diverse than CICERO. We observe this trend among both correct and incorrect answers. As such, \dataset{} provides much richer and diversified multiview commonsense inferences than CICERO. We show a comparative example of annotations in CICERO and \dataset{} in \cref{tab:examples-v1-v2}.

We compute the instance-level average of BLEU~\cite{papineni2002bleu},
% METEOR~\cite{banerjee2005meteor}, 
ROUGE~\cite{lin2004rouge}, CIDEr~\cite{vedantam2015cider}, and semantic similarity among all (correct, correct), (incorrect, incorrect) and (correct, incorrect) answer pairs in \cref{tab:sim}. 
% We report BLEU over uni-grams, bi-grams, four-grams and 
We use the \textit{all-mpnet-base-v2} model
% \footnote{\url{https://huggingface.co/sentence-transformers/all-mpnet-base-v2}}
~\cite{reimers-2019-sentence-bert} to compute the semantic similarity. All scores are reported between 0-1, with a higher score indicating more similarity. The numbers reported in \cref{tab:sim} clearly indicate that answers in CICERO are significantly less diverse. We also conclude that annotations in \dataset{} provide a superior quality of multiview commonsense inferences.
Similar to \cite{ghosal-etal-2022-cicero}, we carry out a quality assurance stage on \dataset{}, details of which can be found in \cref{sec:quality}.

\section{\model{}} \label{sec:pretraining_methods}
We propose DIALogue-level Commonsense Transformer -- \model{}, a pretrained transformer for commonsense inference in dialogues. It is a model trained on a variety of dialogue-related tasks, which help the model better leverage the structural information from the dialogues. The model can be used as the initial weight for further downstream task finetuning and have the capability of making zero-shot inference on its own. 

The pretraining tasks include choosing the correct options illustrated, recovering corrupted input, sorting, and generation based on concepts augmented input. All the pretraining tasks are built based on only the training set of CICERO to avoid information leaking from seeing the dialogues in the test set. We describe details for each training objective in the following sections. 

\subsection{Problem Formulation}
Given a dialogue $\mathcal{D}$ consists of $n$ utterances: $[u_1, u_2, ..., u_n]$, our task is to predict the correct answers from a set of choices for questions on a target utterance $u_i$ in CICERO as illustrated in \cref{sec:dataset}: \textit{cause (c), effect (e), motivation (m), prerequisite (p)}, and \textit{reaction (r)}. We denote the questions as $Q = [Q^j] $, where $j=[0, 1,.., 4]$ corresponds to the relation type being asked. The annotated answers on target utterance $u_i$ are represented as $A_i = [a_i^j] = [c_i, e_i, m_i, p_i, r_i]$, for each aforementioned question type respectively. 
Each $a_i^j$ consists of multiple choices, among which at least two are correct answers. For the pretraining, we use $a_i^j$ to refer to one of the correct answers if not indicated otherwise.
We denote the non-stopword nouns and verbs for either utterance $u_i$ or the corresponding answer as concept $c_i$.
Note that not all five questions are annotated for each target utterance in CICERO. Hence, for a particular target utterance, $A_i$ and $Q$ contain only a subset of the question types, making the value of $j$ no more than four. 

\subsection{Pre-training objectives}
\label{sec:pretrain-objective-groups}
We propose a set of objectives to train the model in a text-to-text manner. The input usually consists of a combination of the prompt text denoting the task referred to as $p$, the concatenation of utterances in the dialogue $\mathcal{D}$ referred to as context $x$, and objective-specific information detailed in the following section. Different parts of the input are concatenated to form the input sequence, separated by special tokens and text indicating the parts. We give details on the input formats and prompts used for all the pre-training objectives in \cref{tab:generations1}.

\subsubsection{Primary Objectives (PO)}
Primary objectives train open-ended text generation without any set of options to choose from, as in the contextual commonsense inference task: 
\begin{enumerate}[label=(\roman*), leftmargin=*, wide, itemsep=0pt, labelwidth=!, labelindent=0pt]
\setlength\itemsep{-0.25em}
    \item Given context $x$, target utterance $u_i$, and question $Q^j$, generate the corresponding answer $a_i^{j}$. 
    \item Given context $x$, question $Q^j$, and its answer $a_i^{j}$, generate the corresponding target utterance $u_i$.
    
    \item Given context $x$, target utterance $u_i$, question $Q^j$, answer $a_i^j$, and question or another type $Q^k$, generate the corresponding answer $a_i^k$.  
\end{enumerate}

\subsubsection{Single Correct Answer Objectives (SCAO)}
These objectives ask the model to generate the right choice from the given options or a closed set of relations, with access to the dialogue context.

\begin{enumerate}[label=(\roman*), leftmargin=*, wide, itemsep=0pt, labelwidth=!, labelindent=0pt]
\setlength\itemsep{-0.25em}
	\item Given context $x$, target utterance $u_i$, question $Q^j$, multiple answer choices $\bar{a}_i^j$, generate the correct answer $a_i^j$. The answer choices $\bar{a}_i^j$ includes correct answer $a_i^j$ and incorrect answers $a_i^{j-}$. We concatenate the question, target, context, and answer choices with separators to form the input.
	\item Given context $x$, target utterance $u_i$, answer $a_i^j$, generate the question type of $Q^j$. We concatenate the answer, target, and context to form the input. The output is one of the five question type strings: \textit{cause, effect, motivation, prerequisite}, or \textit{reaction}.
	\item Given context $x$, answer $a_i^j$, question $Q^j$, a pool of possible target utterances $\bar{u}_i$, choose the correct target utterance $u_i$. The pool includes correct target utterance $u_i$ and three other utterances $u_i^-$ randomly sampled from the same dialogue. 
    % 	We concatenate the question, target, context, and answer choices with separators to form the input.
\end{enumerate}

\subsubsection{Concept-Based Objectives (CO)}
\label{sec:ccr}
These objectives train to reconstruct a sentence from the set of concepts it contains, and generate the answer or target from the concepts in the target or answer, respectively. The concepts are selected based on the part-of-speech tags parsed by \textit{Spacy}\footnote{\url{https://spacy.io/}} after removing stop words. 

\begin{enumerate}[label=(\roman*), leftmargin=*, wide, itemsep=0pt, labelwidth=!, labelindent=0pt]
\setlength\itemsep{-0.25em}
	\item Given context $x$, question $Q^j$, concepts $c_i^j$ from answer $a_i^j$, generate the target utterance $u_i$. We use the concatenation of a template question and context as the input.
	
	\item Given context $x$, concepts $c_i$ from target utterance $u_i$, question $Q^j$, answer $a_i^j$, generate the target utterance $u_i$. Following a strategy similar to the previous case, we use $x, a_i^j, c_i, Q^j$ along with a template question to form the input.
	
	\item Given context $x$, question $Q^j$, and concepts $c_i$ from target utterance $u_i$, generate the answer $a_i^j$. We concatenate the question, concepts, and context to form the input.
	
	\item Given context $x$, target utterance $u_i$, question $Q^j$, and concepts $c_i^j$ from answer $a_i^j$, generate the answer $a_i^j$. We concatenate the question, target, concepts, and context to form the input.
\end{enumerate}

\subsubsection{Denoising Objectives (DO)}
These objectives train to restore and order the corrupted concepts in the target utterance or answer. Corruption is performed by randomly changing the order of the concepts, and randomly removing one concept in the original utterance or answer. A similar concept order recovery has previously been explored by~\citet{zhou2021pre}.
\begin{enumerate}[label=(\roman*), leftmargin=*, wide, itemsep=0pt, labelwidth=!, labelindent=0pt]
\setlength\itemsep{-0.25em}
	\item Given context $x$, target utterance $u_i$, question $Q^j$, corrupted concepts $\bar{c}_i^j$ for answer $a_i^j$, generate correct concepts $c_i^j$.

	\item Given context $x$, question $Q^j$, answer $a_i^j$, corrupted concepts $\bar{c}_i$ for utterance $u_i$, generate correct concepts $c_i$.
\end{enumerate}

\subsubsection{Sorting-Based Objectives (SO)}
% (related to the dialogue structure, o10 matches answer with question 
Sorting-based objectives require the model to be aware of the order of the utterances and the order of questions asked in the dialogue.
\begin{enumerate}[label=(\roman*), leftmargin=*, wide, itemsep=0pt, labelwidth=!, labelindent=0pt]
\setlength\itemsep{-0.25em}
	\item We consider the following precedence order of the relations: $c \rightarrow p \rightarrow m \rightarrow e \rightarrow r$. Now, given context $x$ and a randomly ordered subset of answers $\hat{a}$ from $A$, the objective is to generate the sorted order of $\hat{a}$ according to utterance location and relation precedence. The output to be generated is formulated according to indices of answers in the subset $\hat{a}$. For instance, if $\hat{a} = [r_5, p_2, e_0, c_0, m_2]$, the output to generate would be \texttt{3 2 1 4 0}, denoting the sorted order $c_0 \rightarrow e_0 \rightarrow p_2 \rightarrow m_2 \rightarrow r_5$.
	\item Given a randomly ordered set of utterances $\hat{u}$ from $\mathcal{D}$, identify the correct order. For example if $\hat{u} = [u_3, u_1, u_2]$, the output to be predicted is the string \texttt{1 2 0}, assuming indexing starts from 0.
\end{enumerate}	

\section{Experiments} \label{sec:experiments}

We evaluate the effectiveness of \model{} on commonsense inference tasks with the multi-choice question answering (MCQ) format under various settings, where we compare the performance of models finetuned on the MCQ task based on \model{} with the baselines.

\subsection{Experimental Setup} \label{subsec:pretraining_setup}
\paragraph{Pretraining.}
\looseness=-1 We pretrain \model{} using T5-large as the backbone (770M parameters). We initialize the parameters with the checkpoint released by \cite{raffel2019exploring} and continue pretraining in a text-to-text manner instead of span filling. We use the Adafactor \cite{shazeer2018adafactor} optimizer with a weight decay of 0.005 and a learning rate of 1e-5. Note that Adafactor significantly reduces the memory footprint for conversational tasks with long text input. We train the model for 75000 steps with a batch size of 16. The training takes around 22 hours on two A40 GPUs.

\paragraph{Finetuning.}
We finetuned the model based on either \model{} or T5-large. We use the Adafactor optimizer during pretraining with a learning rate of 3e-5. All finetuning experiments are run for 5 epochs with five different random seeds. Each trial takes 30 minutes on an A40 GPU. 

\paragraph{Evaluation Metrics}
We use macro-F1 and Exact Match to evaluate the performance of the models.

\begin{table}[t]
%\small
\centering
\resizebox{\linewidth}{!}{
\begin{tabular}{l|c|c|cccccc}
\toprule
\multirow{2}{*}{\textbf{Model}} & \multirow{2}{*}{\textbf{Finetuned on}} & \multirow{2}{*}{\textbf{\begin{tabular}[c]{@{}c@{}}Avg\\ Macro F1\end{tabular}}} & \multicolumn{6}{c}{\textbf{Exact Match}}\\
&&  & \textbf{Cause} & \textbf{Subseq} & \textbf{Prereq} & \textbf{Motiv} & \textbf{Reaction} & \textbf{Average} \\
\midrule
\multirow{1}{*}{T5$_{Large}$}  & CICERO    & 0.7001 & 0.2521 & 0.2358 & 0.2430 & 0.3258 & 0.3258 & 0.2566\\

\multirow{1}{*}{\model{}$_{Large}$}  & CICERO  & \bf 0.7066 & \bf 0.2736 & \bf 0.2560 & \bf 0.2457 & \bf 0.3539 &\textbf{0.3420} & \bf 0.2754\\
\midrule
\multirow{1}{*}{T5$_{Large}$} & \dataset{} & 0.8795 & 0.6552 & 0.7148 & - & \bf 0.7587 & 0.7243 & 0.7195 \\
\multirow{1}{*}{\model{}$_{Large}$} & \dataset{} & \bf0.8863 & \bf 0.6905 & \bf 0.7388 & - & 0.7537 & \bf 0.7614 & \bf 0.7380\\
\bottomrule
\end{tabular}
}
\caption{\footnotesize Performance of \model{} on CICERO and \dataset{}.}
\vspace{-0.30mm}
\label{tab:main_results}
\end{table}
\subsection{Overall Results on CICERO} \label{subsec:v1v1}
We evaluate \model{} with MCQ from the CICERO dataset it pretrained on. 
Table \ref{tab:main_results} shows the performance on the MCQ task. 
We find that \model{} improves the performance compared to the baseline on all metrics except recall. For the exact match, there is around 2\% universal improvement for all inference types, indicating that the pretraining is not limited to a certain type of commonsense inference. 
The results suggest that, although having the same access to dialogue context and question-answer pairs from the same dataset, the pretraining helps exploit the information in the dataset. 

\subsection{Transferability of Pretraining} \label{subsec:v1v2}
To further investigate if the performance boost comes from merely seeing the questions and choices in advance. We test \model{} on newly collected \dataset{}.  
Table \ref{tab:main_results} shows that \model{} outperform the T5-large baseline on all metrics again. There is a similar trend of improvement across inference types for exact matches. The results show information learned in \model{} generalize to MCQ samples drawn from a different distribution. Interestingly, despite seeing answers of CICERO during pre-training, the performance of \model{} on CICERO is worse than its performance on \dataset{}. We think this could be due to the high lexical overlap and semantic similarity between correct and incorrect answers in CICERO (as shown in \cref{tab:sim}) that might cause confusion in easily finding the decision boundary. As a result,  both T5-large and \model{} perform poorly to predict multiple correct answers in CICERO.

\begin{table}[t]
%\small
\centering
\resizebox{\linewidth}{!}{
\begin{tabular}{l|c|cccccc}
\toprule
\multirow{2}{*}{\textbf{Objectives}} & \multirow{2}{*}{\textbf{\begin{tabular}[c]{@{}c@{}}Avg\\ Macro F1\end{tabular}}} & \multicolumn{6}{c}{\textbf{Exact Match}}\\
% & Pre & Rec & F1 & Pre & Rec & F1 & Cause & Subseq & Prereq & Motiv & Reaction & Avg\\
& & \textbf{Cause} & \textbf{Subseq} & \textbf{Prereq} & \textbf{Motiv} & \textbf{Reaction} & \textbf{Average} \\
\midrule
\multirow{1}{*}{T5-Large} & 0.7001 & 0.2521 & 0.2358 & 0.2430 & 0.3258 & 0.3258 & 0.2566\\
% \multirow{1}{*}{ASER-only}      &0.7116 & 0.6333 & 0.6706 & 0.7172 & 0.6501 & 0.682 & 0.249 & 0.238 & 0.2364 & 0.3127 & 0.3103 & 0.2534 \\
\multirow{1}{*}{All}        & 0.7066 & 0.2736 & 0.2560 & 0.2457 & 0.3539 &\textbf{0.3420} & 0.2754\\
% \multirow{1}{*}{+Dialog Act}    & 0.7302 & 0.6568 & 0.6916 & 0.7372 & 0.6787 & 0.7068 & 0.2815 & 0.2545 & 0.2536 & 0.3471 & 0.2971 & 0.2763\\
% \multirow{1}{*}{+Dialog Act2}    & 0.7311 & 0.6581 & 0.6927 & 0.7389 & 0.6805 & 0.7085 & 0.2918 & 0.2530 & 0.2656 & 0.3557 & 0.2931 & 0.2790 \\
% \multirow{1}{*}{+ASER}    & 0.7303 & 0.6545 & 0.6903 & 0.7371 & 0.6756 & 0.705 & 0.2847 & 0.2481 & 0.2629 & 0.3471 & 0.2988 & 0.2741 \\
\multirow{1}{*}{-PO}     & 0.7069 & \textbf{0.2964} & 0.2534 & \bf 0.2722 & \textbf{0.3660} & 0.3190 & \textbf{0.2841}\\
\multirow{1}{*}{-SCAO}        & 0.6963 & 0.2613 & \textbf{0.2797} & 0.2324 & 0.3419 & 0.3276 & 0.269\\
\multirow{1}{*}{-CO}       & \textbf{0.7096} & 0.2867 & 0.2587 & 0.2563 & 0.3505 & 0.3276 & 0.2803\\
\multirow{1}{*}{-DO}  & 0.7036 & 0.2737 & 0.2530 & 0.2430 & 0.3505 & 0.2931 & 0.2703\\
\multirow{1}{*}{-SO}      & 0.7090 & 0.2834 & 0.2609 & 0.2656 & 0.3626 & 0.3074 & 0.2815 \\
\midrule
\multirow{1}{*}{Ensemble} & - & 0.2964 & 0.2797 & 0.2722 & 0.3660 & 0.3420 & 0.3112\\
\bottomrule
\end{tabular}
}
\caption{\footnotesize Ablation study on CICERO. Reported results are the average of five different runs. The Ensemble model selects the best-performing ablated model for a particular relation.}
\label{tab:ablation_results_v1}
\end{table}

\begin{table}[h]
%\small
\centering
\resizebox{\linewidth}{!}{
\begin{tabular}{l|c|ccccc}
\toprule
\multirow{2}{*}{\textbf{Objectives}} & \multirow{2}{*}{\textbf{\begin{tabular}[c]{@{}c@{}}Avg\\ Macro F1\end{tabular}}} & \multicolumn{5}{c}{\textbf{Exact Match}}\\
% & Pre & Rec & F1 & Pre & Rec & F1 & Cause & Subseq & Motiv & Reaction & Avg\\
&  & \textbf{Cause} & \textbf{Subseq} & \textbf{Motiv} & \textbf{Reaction} & \textbf{Average} \\
\midrule
\multirow{1}{*}{T5-Large} & 0.8795 & 0.6552 & 0.7148 & 0.7587 & 0.7243 & 0.7195 \\
\multirow{1}{*}{All}& 0.8863 & 0.6905 & 0.7388 & 0.7537 & 0.7614 & 0.7380\\
% \multirow{1}{*}{+Dialog Act} & 0.8913 & 0.8882 & 0.8897 & 0.8852 & 0.8824 & 0.8838 & 0.6982 & 0.7286 & 0.7446 & 0.7714 & 0.7324 \\
\multirow{1}{*}{-PO} & 0.8854 & \textbf{0.7105} & 0.7303 & 0.7480 & 0.7619 & 0.7354\\
\multirow{1}{*}{-SCAO} & 0.8840 & 0.6790 & 0.7362 & 0.7516 & 0.7333 & 0.7320 \\
\multirow{1}{*}{-CO} & \textbf{0.8900} & 0.7065 & \textbf{0.7408} & 0.7613 & 0.7714 & \textbf{0.7443}\\
\multirow{1}{*}{-DO}& 0.8866 & 0.7023 & 0.7315 & 0.7550 & \textbf{0.7738} & 0.7378\\
\multirow{1}{*}{-SO}& 0.8867 & 0.6872 & 0.7273 & \textbf{0.7662} & 0.7666 & 0.7360\\
\midrule
\multirow{1}{*}{Ensemble} & - & 0.7105 & 0.7408 & 0.7662 & 0.7738 & 0.7478 \\
\bottomrule
\end{tabular}
}
\caption{\footnotesize Ablation study on \dataset{}. Reported results are the average of five different runs. %The Ensemble model selects the best-performing ablated model for a particular relation.
}
\label{tab:ablation_results_v2}
\end{table}

\subsection{Ablation Study of Pretraining Objectives} \label{subsec:ablation}
% We conduct ablation studies on different groups of pretraining objectives on both CICERO and \dataset{}. 
For a fair comparison, we remove a group of pretraining objectives for each setting and pretrain the model with the exact same set of hyper-parameters, including the random seeds, all for five epochs.
Table \ref{tab:ablation_results_v1} shows that all the ablation models still outperform the baseline, meaning that there is at least more than one group of helpful objectives. Removing the \textit{Single Correct Answer Objectives} i.e., SCAO causes the largest drop among all metrics, suggesting it carries essential information. 
On contrary, removing \textit{Primary Objectives} and \textit{Sorting Based Objectives} leads to slightly higher metrics. One plausible explanation is that the gap in the input format for these objectives causes trouble for later finetuning. For example, \textit{Sorting Objectives} ask the model to predict a sequence of integers, which may be confused with the multiple-choice marker. 
The results for \dataset{} is shown in Table \ref{tab:ablation_results_v2}. It holds the same conclusion that all ablation models perform better than the baseline. It is also interesting that the \textit{Concept Objective} i.e., CO ablation group gets the highest performance on most of the metrics, suggesting that the concepts from CICERO may misalign with the ones in \dataset{}.

\subsection{Performance Analysis} \label{subsec:analysis}
\paragraph{Cross-Dataset Performance.}

\begin{table}[t]
\small
\newcommand*{\gr}[1]{\textcolor{darkgreen}{#1}}
\newcommand*{\rd}[1]{\textcolor{red}{#1}}
\centering
\resizebox{0.9\linewidth}{!}{
\begin{tabular}{l|c|cccccc}
\toprule
\multirow{2}{*}{\textbf{Train}} & \multirow{2}{*}{\textbf{Test}} & \multicolumn{6}{c}{\textbf{Exact Match}}\\
% & Pre & Rec & F1 & Pre & Rec & F1 & Cause & Subseq & Prereq & Motiv & Reaction & Avg\\
& & \textbf{Cause} & \textbf{Subseq} & \textbf{Prereq} & \textbf{Motiv} & \textbf{Reaction} & \textbf{Average} \\
\midrule
v1-four & \multirow{2}{*}{v1-four} & 0.3307 & 0.3254 & 0.2948 & 0.4794 & 0.431 & 0.3457\\
v2-four & & 0.2451 & 0.253 & 0.2669 & 0.3557 & 0.3448 & 0.2694\\
\multicolumn{2}{c}{$\Delta$} & \gr{0.0856} & \gr{0.0724} & 0.0279 & \gr{0.1237} & \gr{0.0862} & \gr{0.0763} \\
\midrule
v2-four & \multirow{2}{*}{v2-four} & 0.5934 & 0.5858 & - & 0.7244 & 0.6214 & 0.6302\\
v1-four & & 0.1203 & 0.3062 & - & 0.4948 & 0.2857 & 0.3321\\
\multicolumn{2}{c}{$\Delta$} & \rd{0.4731} & \rd{0.2796} & - & \rd{0.2296} & \rd{0.3357} & \rd{0.2981}\\
\bottomrule
\end{tabular}
}
\caption{\footnotesize Cross-dataset results (avg. of five runs) of \model{}; v1-four and v2-four stand for CICERO and \dataset{}, respectively, culled to have four options per sample.}
\label{tab:cross-dataset-perf}
\end{table}

\looseness=-1 \cref{tab:cross-dataset-perf} shows cross-dataset adaptability of \model{}. To circumvent the influence of the variability of answer counts in CICERO and \dataset{}, we cull the samples of both datasets to have exactly four answers. For each sample with more than four answers, two correct answers are randomly picked without replacement, and then two more answers are randomly chosen from the rest. This results in at least two correct answers per sample. As expected, both cross-dataset transfers lead to diminished performance due to the starker difference in distribution between training and test set. Interestingly, the performance drop of 29.81\% for CICERO to \dataset{} transfer is far more severe than the drop of 7.63\% for \dataset{} to CICERO transfer. This observation strongly implies that \dataset{} allows for a much more robust cross-dataset transfer than CICERO. This is likely a consequence of the larger diversity of answers in the training samples of \dataset{}, as indicated in \cref{sec:diversity}. Another observation is the performance improvement (7.03\%) and degradation (10.78\%) on in-dataset transfer for culled CICERO and \dataset{}, respectively. This is indicative of the strong influence of negative samples over the overall performance of \model{} on both datasets.

\begin{figure}[t]
    \centering
    \includegraphics[width=\linewidth]{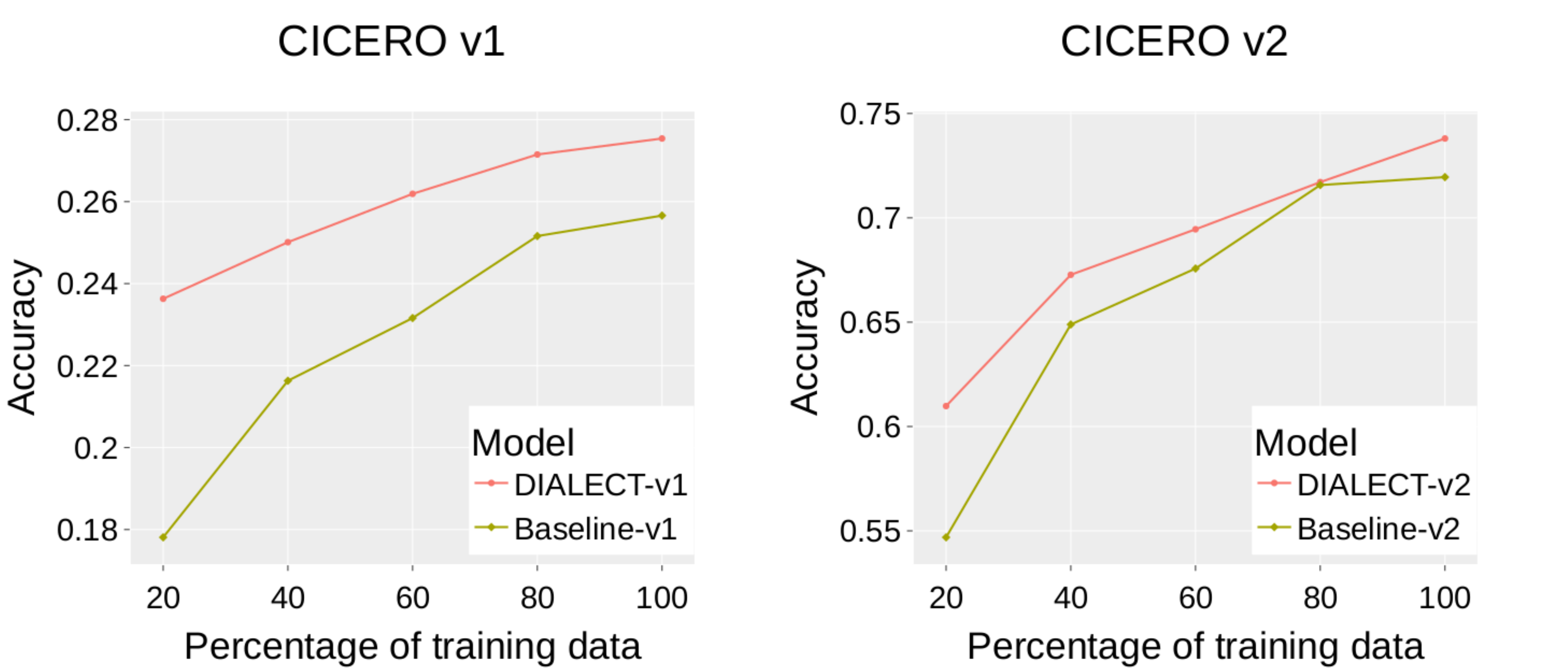}
    \caption{\footnotesize The performance of finetuned models trained with fewer samples.}
    \label{fig:low-resource}
\end{figure}

\paragraph{Performance with Fewer Training Examples.} 
\looseness=-1  To assess our proposed objectives' efficacy in the low-resource setting, we compare the fine-tuning performance of \model{} with T5-Large using different fragments of the training data. As shown in \cref{fig:low-resource}, \model{} consistently attains better exact match accuracy than the T5-Large baseline on both CICERO and \dataset{}. It can be seen that the performance improvement by \model{} is more significant under the low resource setting. When finetuned with 20\% of the training data, \model{} offers over 5\% performance boost on both datasets, compared with around 2\% for the full dataset. 
This indicates that \model{} might be endowed with some commonsense knowledge through its pre-training using our proposed objectives. 
As a result, \model{} does not require much training data before attaining a decent performance. Note that, although building the pre-training objectives relies on the training set of CICERO, the training set of \dataset{} is not used, and thus can be considered as a "true low resource setting".
In contrast, the baseline T5-Large model needs more training data before obtaining a good fine-tuning performance. Based on these observations, we may conjecture that T5-Large lacks the required commonsense knowledge that \model{} encodes in its parameters.

\section{Related Works}
The area of commonsense reasoning has received significant attention recently, with the introduction of several new benchmarks~\cite{zellers-etal-2018-swag,talmor2019commonsenseqa,bisk2020piqa}. The benchmarks target evaluation of commonsense across various dimensions - causality~\cite{roemmele2011choice}, social commonsense with question answering~\cite{sap2019social}, abduction~\cite{bhagavatula2020abductive}, etc. Language models specifically trained for commonsense reasoning across these dimensions have also been proposed~\cite{lourie2021unicorn}. Despite the progress in those directions, the multiview aspect of question answering and commonsense reasoning, particularly, has been an under-explored area. Recently, ~\citet{zhu-etal-2020-question} proposed a dataset for extractive multiple-span question answering; ~\citet{qin2021timedial} introduced a dataset for temporal reasoning in dialogues with multiple correct answers satisfying certain temporal properties. ~\citet{ghosal-etal-2022-cicero} introduced the CICERO dataset for dialogue reasoning with contextual commonsense inference. Although the dataset contains multiple speculative inferences, there are certain shortcomings as inferences are less diverse, and not all inferences are human-written. We motivate our work against this aspect and present a dataset with significantly richer and more diverse multiview commonsense inferences from dialogues, all of which are human-written.

\section{Conclusion}
We introduce \dataset{}, a human-written dataset for distinct multiview commonsense inferences in dialogues. The dataset contains $\sim$8.3k instances from $\sim$2.3k dialogues across four commonsense dimensions -- cause, subsequent event, motivation, and reaction.
We also propose \model{}, which is pre-trained on a collection of dialogue understanding objectives. We evaluate it on the multiview commonsense inference task and analyze its performance across various settings.

\section{Limitations}
Our model \model{} can only perform the answer selection task (MCQ). Wherein the commonsense inference generation as proposed in \cite{ghosal-etal-2022-cicero} is more challenging which \model{} can not solve. Besides, \model{} requires heavy computing power for pre-training and fine-tuning. As a consequence, it can not be deployed on mobile devices with very low computational power. On the other hand, our proposed \dataset{} only contains inferences across four different commonsense dimensions -- cause, subsequent event, motivation, and reaction. Hence, models e.g., \model{} trained on \dataset{} could be limited in their capacity to infer other types of commonsense relations.

\section*{Ethics Statement}
The annotators for \dataset{} were hired through a data annotation service. The compensation was derived based on the country of residence of the annotators, as deemed by the company. The study has been categorized as ``exempt'' by the IRB. Annotators were strictly asked not to write any toxic content (hateful or offensive toward any gender, race, sex, or religion). They were asked to consider gender-neutral settings in dialogues whenever possible.

The source dialogue datasets -- DailyDialog, MuTual, and DREAM are high-quality multi-turn dialogue datasets manually annotated by experts in dialogue, communication theory, and linguistics. All three datasets have been extensively used and studied in the natural language processing literature. The three source datasets and our annotations in \dataset{} do not contain any personal data or any information that can uniquely identify individual people or groups.

\bibliography{refs}
\bibliographystyle{acl_natbib}
% \newpage
\clearpage
\appendix
\section{Annotation of Emotional Reaction}
The annotators capture the appropriate emotion of the listener using the emotion terms listed in \cref{tab:emotions} using verbatim or related words, to write the answer for the question \textit{What is the possible emotional reaction of the listener: A (or B)?}

\section{Quality Assurance of \dataset{}}
\label{sec:quality}
The dataset quality is ensured with the following steps:
\begin{itemize}[leftmargin=*, wide, itemsep=0em, labelwidth=!, labelindent=0pt]
\setlength\itemsep{-0.25em}
\item Initially, we sample $30$ random dialogues and manually annotate all the questions in those. Each annotator is then evaluated on those dialogues and is selected for the annotation task if 95\% of his/her annotations are approved by us.  
\item We constantly review and provide feedback to the annotators during the annotation process. Annotators are also instructed to amend their answers.
\item Upon completion of the annotation, we employ three additional annotators who manually check the annotated samples and score  their acceptability. These annotators reached a consensus for approving 96.2\% of these samples. The samples not bearing majority agreement were removed from the dataset. The statistics of the annotated dataset are shown in \cref{tab:stat}. A number of annotated examples from \dataset{} are also shown in \cref{tab:examples-v1-v2}.
\end{itemize}

% \begin{table}[t]
% \centering
% \resizebox{\linewidth}{!}{
% 	\begin{tabular}{r@{~~}|r@{~~}|r@{~~}|r@{~~}|r@{~~}}
% 	\toprule
% 	Admiration & Affection & Afraid & Angry & Annoyed \\
%     Anticipating & Anxious & Apprehensive & Ashamed & Awe \\
%     Awkwardness & Boredom & Calmness & Caring & Confident \\
%     Confusion & Content & Craving & Devastated & Disappointed \\
%     Disgusted & Eagerness & Embarrassed & Encouragement & Enthusiasm \\
%     Excited & Faithful & Fear & Furious & Grateful \\
%     Gratitude & Guilty & Happy & Hopeful & Impressed \\
%     Interest & Jealous & Joyful & Lonely & Nostalgic \\
%     Prepared & Proud & Relief & Romance & Sad \\
%     Satisfaction & Sentimental & Surprised & Terrified & Trusting \\
%     \bottomrule
% 	\end{tabular}
% 	}
% 	\caption{Possible emotional reactions of the listener.}
% 	\label{tab:emotions}
% \end{table}

\begin{table}[t]
\centering
\resizebox{\linewidth}{!}{
	\begin{tabular}{r@{~~}|r@{~~}|r@{~~}|r@{~~}}
	\toprule
	Admiration & Affection & Afraid & Angry \\
	Annoyed & Anticipating & Anxious & Apprehensive \\ 
	Ashamed & Awe & Awkwardness & Boredom \\
	Calmness & Caring & Confident & Confusion \\
	Content & Craving & Devastated & Disappointed \\
    Disgusted & Eagerness & Embarrassed & Encouragement \\
    Enthusiasm & Excited & Faithful & Fear \\
    Furious & Grateful& Gratitude & Guilty \\ 
    Happy & Hopeful & Impressed & Interest \\
    Jealous & Joyful & Lonely & Nostalgic \\
    Prepared & Proud & Relief & Romance \\
    Sad & Satisfaction & Sentimental & Surprised \\
    Terrified & Trusting \\
    \bottomrule
	\end{tabular}
	}
	\caption{List of possible emotional reactions of the listener.}
	\label{tab:emotions}
\end{table}
\begin{table}[t]
\small
\centering
\resizebox{\linewidth}{!}{
	\begin{tabular}{p{4.5cm}@{}|c@{~~}|c@{~~}}
	\toprule
	\textbf{Description} & \textbf{\# Instances} & \textbf{Percentage}\\
	\midrule
	\bf \# Dialogues / \# Inferences & & \\
    $\quad$ DailyDialog & 3,280 / 30,509 & 57.82 / 57.34 \\
    $\quad$ MuTual & 1,640 / 14,207 & 28.91 / 26.70 \\
    $\quad$ DREAM & 753 / 8,488 & 13.27 / 15.95\\
    $\quad$ \bf Total & 5,673 / 53,204  & -- \\
    \midrule
    \# \bf Dialogues with \# Inferences & & \\
    $\quad$ less than 10 & 3,140 & 55.35 \\
    $\quad$ between 10-20 & 2,518 & 44.39 \\
    $\quad$ between 21-30 & 15 & 0.26\\
    \bf Avg. \# Inferences per Dialogue & 9.38 & --\\
    \midrule
    \begin{tabular}{l}
    \bf Instances with\\ \bf \# Correct Answers \end{tabular} & & \\
    $\quad$ only 1 & 45759 & 86.01 \\
    $\quad$ only 2 & 4985 & 9.37 \\
    $\quad$ $>$ 2 & 2460 & 4.62 \\
    \midrule
    \begin{tabular}{l}
    \bf Inference Types in \\ \bf Train / Validation / Test \end{tabular} & &  \\
    $\quad$ Cause & 10,386 / 3,060 / 3,071 & 33.06 / 28.10 / 28.18 \\
    $\quad$ Subsequent Event & 6,617 / 4,021 / 4,050 & 21.06 / 36.93 / 37.16 \\
    $\quad$ Prerequisite & 7,501 / 1,347 / 1,396 & 23.87 / 12.37 / 12.81 \\
    $\quad$ Motivation & 4,412 / 1,420 / 1,401 & 14.04 / 13.04 / 12.86 \\
    $\quad$ Reaction & 2,502 / 1,040 / 980\hspace{0.2cm} & 7.96\hspace{0.1cm}
    /\hspace{0.17cm}9.55\hspace{0.17cm}/\hspace{0.1cm} 8.99\\
    \bottomrule
	\end{tabular}
	}
	\caption{Statistics of CICERO~\cite{ghosal-etal-2022-cicero}.}
	\label{tab:cicerostat}
\end{table}
\begin{table}[t]
\small
\centering
\resizebox{0.95\linewidth}{!}{
\begin{tabular}{lccC{1.5cm}}
\toprule
\multirow{2}{*}{\textbf{Group}} & \multirow{2}{*}{\textbf{\# Instances}} & \multirow{2}{*}{\textbf{Sub-group}} & \textbf{Sub-group \# Instances} \\
\midrule

\multirow{2}{*}{\textbf{PO}} & \multirow{2}{*}{107,198} & (i), (ii) & 31,418 \\
& & (iii) & 44,362 \\
\midrule 
\textbf{SCAO} & 94,254 & (i) - (iii) & 31,418 \\
\midrule 
\textbf{CO} & 125,672 & (i) - (iv) & 31,418 \\
\midrule 
\multirow{2}{*}{\textbf{DO}} & \multirow{2}{*}{60,302} & (i) & 31,369 \\
 & & (ii) & 28,933 \\
\midrule 
\multirow{2}{*}{\textbf{SO}} & \multirow{2}{*}{6,953} & (i) & 3,476 \\
 & & (ii) & 3,477 \\
\midrule
\textbf{Total} & 394,379 & - & - \\ 
\bottomrule
\end{tabular}
}
\caption{The number of instances for each group and corresponding sub-groups of objective functions as described in \cref{sec:pretrain-objective-groups}. The number of training instances in CICERO is 31,418, which is also the number of instances in some of the sub-groups.}
\label{tab:pretrain-stat}
\end{table}

\begin{table*}[t]
  \centering
  \begin{subtable}{\linewidth}
 %  \small
  \centering
  \resizebox{\linewidth}{!}{
    \begin{tabular}{p{15cm}}
    \toprule
      \textbf{A \pmb {($u_1$)}}: What's that smell?
      \textbf{A \pmb {($u_2$)}}: Are you making a chocolate cake?
      \textbf{A \pmb {($u_3$)}}: I smell something different, peers?
      \textbf{B \pmb {($u_4$)}}: No, I'm making chocolate banana cookies.
      \textbf{B \pmb {($u_5$)}}: At first I was going to use the oranges , but I think these will taste better.
      \\
      \toprule
      \textbf{Target -} \pmb{$u_4$}; \textbf{Question:} \colorbox{Green2}{\bf Motivation}; \textbf{Correct Answers in \dataset{}:}  i) The speaker has leftover chocolate and bananas and wants to consume them quickly. ii) The speaker likes chocolate sweets. \textbf{Incorrect Answers in \dataset{}:} i) The speaker wants to make the kitchen smelly to stop the listener entering. ii) The speaker is hungry and chocolate is not filling enough. \\
      \midrule
      \textbf{Target -} \pmb{$u_4$}; \textbf{Question:} \colorbox{ghostwhite}{\bf Subsequent Event}; \textbf{Correct Answers in CICERO:}  i) The listener will request his friend to taste the cookies he prepared just now. \textbf{Correct Answers in \dataset{}:} i) The speaker asks the listener to pass the spatula to her.  \textbf{Incorrect Answers in CICERO:} i) The listener will ask his friends to taste the cake he prepared just now. ii) The listener will request his friends to taste the chocolate cake he prepared just now. \textbf{Incorrect Answers \dataset{}:} i) The speaker invites the speaker to taste the orange cookies. ii) The listener asks the speaker to get out of the kitchen then takes over the cookies.\\
      \midrule
      \textbf{Target -} \pmb{$u_5$}; \textbf{Question:} \colorbox{brilliantlavender}{\bf Cause}; \textbf{Correct Answers in CICERO:}  i) The speaker was making banana cookies. \textbf{Correct Answers in \dataset{}:} i) It is too difficult to process the orange pulp. ii) The orange smell doesn't match well with chocolate.  \textbf{Incorrect Answers in CICERO:} i) The speaker is making a chocolate cake. ii) The speaker was baking a cake. \textbf{Incorrect Answers in \dataset{}:} i) The orange smell matches much better with chocolate compared with banana. ii) The speaker loves the taste of orange and the texture of its pulp.\\
      \midrule
      \textbf{Target -} \pmb{$u_5$}; \textbf{Question:} \colorbox{unitednationsblue}{\bf Emotional Reaction}; \textbf{Correct Answers in CICERO:} i) The listener is excited to eat the cookies. \textbf{Correct Answers in \dataset{}:} i) The listener feels pity that she cannot have orange cookies. \textbf{Incorrect Answers in CICERO:} i) The listener is excited eats the salad. ii) The listener is excited to eat the muffins instead. \textbf{Incorrect Answers in \dataset{}:} i) The listener is happy to taste orange cookies. ii) The listener is annoyed by the banana smell. \\
      \bottomrule
  \end{tabular}
  }
\end{subtable}
%\vspace{-1.1em}
\caption{ Annotated examples in CICERO and \dataset{} marked with the target utterance and the question type. The first (\textit{dialogue}, \textit{target}, \textit{question}) instance is not present in CICERO. We show the incorrect answers in CICERO for the other three instances. For these instances, the first correct answer is the primary human written answer in CICERO. Incorrect answers in CICERO are significantly less diverse than \dataset{}.}
\label{tab:examples-v1-v2}
\end{table*}

\section{Additional Details on the Pre-training}

We use the CICERO dataset to pre-train \model{}. Detailed statistics of this CICERO dataset are presented in \cref{tab:cicerostat}. The objective-wise statistics of the training dataset used in pre-training \model{} are reported in \cref{tab:pretrain-stat}.

\section{Annotation Details}
We recruited 32 student helpers who are undergraduate students studying computer science and fluent in speaking and writing English. These students have knowledge of Artificial Intelligence. The annotators were paid 7.5 USD per hour which is a standard rate for hiring student helpers at our university. In total, the total cost of the annotation was 2955 USD.

\begin{figure}[h!]
    \centering
    \includegraphics[width=\linewidth]{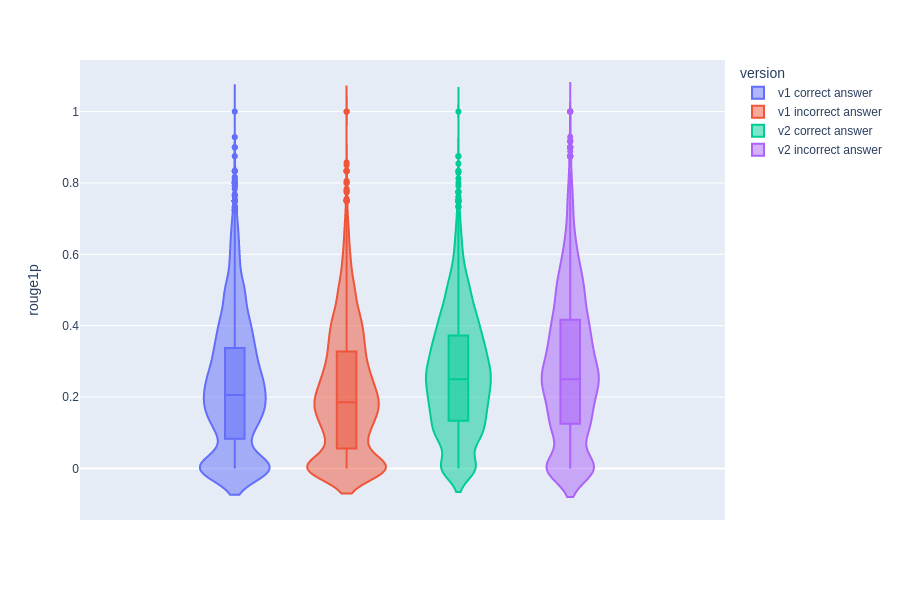}
    \caption{The distribution of Rouge1-P for correct/incorrect answers in CICERO and \dataset{}. CICERO has more samples with a zero ROUGE score, and both correct and incorrect answers from \dataset{} have a slightly higher average ROUGE score than its counterpart.}
    \label{fig:rouge1p_distribution}
\end{figure}

\section{Additional Performance Analysis}

\paragraph{Impact of Lexical Overlap of Answers and Context.}
We use ROUGE-1 precision as the lexical similarity measure between answers and context. For each sample in the training set, we calculate the average ROUGE score for its answers having dialogue context as the reference of this calculation. The distribution of ROUGE scores is shown in \cref{fig:rouge1p_distribution}.

\begin{table*}[t]
%\small
% \newcommand*{\gr}[1]{\textcolor{darkgreen}{#1}}
% \newcommand*{\rd}[1]{\textcolor{red}{#1}}
\centering
\resizebox{\linewidth}{!}{
\begin{tabular}{l|c|c|c|c|cccccc}
\toprule

\multirow{2}{*}{\textbf{Train}} & \multirow{2}{*}{\textbf{Test}} & \multirow{2}{*}{\textbf{R$_c$}} & \multirow{2}{*}{\textbf{R$_i$}} & \multirow{2}{*}{\textbf{|R$_c$}-\textbf{R$_i$}|} &\multicolumn{6}{c}{\textbf{Average Exact Match}}\\
% & Pre & Rec & F1 & Pre & Rec & F1 & Cause & Subseq & Prereq & Motiv & Reaction & Avg\\
& & & & & \textbf{All} & \textbf{All - PO} & \textbf{All - SCAO} & \textbf{All - CO} & \textbf{All - DO} & \textbf{All - SO} 
\\
\midrule
\multirow{2}{*}{CICERO} & CICERO$_{lowr}$ & 0.0103 & 0.1970 & 0.1867 & 23.87 & 26.66 & 24.51 & 25.37 & 22.79 & 25.80\\
& CICERO$_{highr}$ & 0.4715 & 0.2062 & \bf 0.2653 & \bf 28.18 & \bf 30.14 & \bf 26.47 & \bf 29.41 & \bf 30.39  & \bf 28.92\\
\midrule
\multirow{2}{*}{\dataset{}} & \dataset{}$_{-lowr}$ & 0.0665 & 0.1519 & \bf 0.0854 & \bf 75.36 & \bf 72.14 & \bf 72.86 & \bf 73.93 & \bf 74.64 & \bf 75.36\\
& \dataset{}$_{-highr}$ & 0.4961 & 0.4488 & 0.0473 & 72.30 & 71.51 & 70.53 & 73.08 & 71.12 & 70.92\\

\bottomrule
\end{tabular}
}

\caption{The lexical similarity between answers and context also impact models' performance significantly. \textit{lowr} and \textit{highr} denote low rouge precision and high rouge precision groups respectively. R$_c$ and R$_i$ denote the average ROUGE score of correct and incorrect answers, respectively.}
\label{tab:lexical_similarity}
\end{table*}

\begin{figure*}[t]
    \centering
    \includegraphics[width=\linewidth]{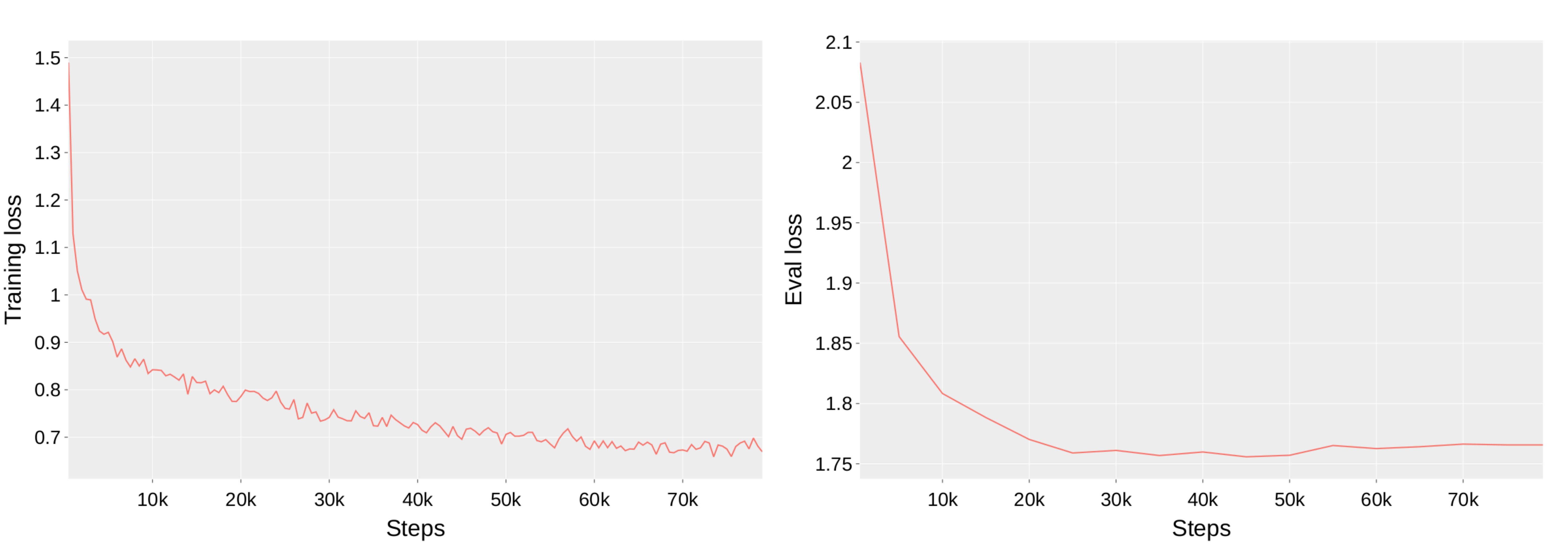}
    \caption{Learning Curve of the Pretraining. The validation loss plateaus and starts to increase after 50k. The best-pretrained model is selected based on its performance on the downstream task of multiple-answer selection.}
    \label{fig:pretrain_learning_curve}
\end{figure*}

We set the lower quartile of the average ROUGE score of training samples as the low threshold and the upper quartile as the high threshold. Based on the two thresholds, we then filter the samples in the test set into low-ROUGE and high-ROUGE groups. 
% Intuitively, a sample within the high-ROUGE group tends to have correct answers that have more overlap with the context. 

\cref{tab:lexical_similarity} shows that the models perform better on the high-ROUGE group of CICERO. This follows the intuition that samples from the high-ROUGE group are easier to predict as they overlap more with the context. 

That is not the case for \dataset{}, where all models perform better in the low-ROUGE group. Upon deeper inspection, we find that the models' performance might be influenced by the gap between the overlap, quantified by the ROUGE score, of the correct ($R_c$) and incorrect answers ($R_i$), instead of the absolute overlap of the correct answers. For the high-ROUGE group of \dataset{}, the incorrect answers have an average ROUGE score of 0.45, which is very close to the score of 0.49 for the correct answers. That may make the separation between the correct and incorrect answers difficult. 

Note that the incorrect answers in CICERO have almost the same average ROUGE scores across both groups, as they are generated automatically.  The correct answers in the high-ROUGE group of CICERO have an average ROUGE score of 0.47, and the incorrect answers in the same group have an average ROUGE score of 0.20. On the other hand, in the low-ROUGE group, correct and incorrect answers have an average ROUGE score of 0.01 and 0.19, respectively. We surmise, as compared to the lower-ROUGE group, the larger gap between the ROUGE scores of the correct and incorrect answers in the high-ROUGE group of CICERO aids \model{} to attain better performance in this group.

\paragraph{Pre-training Steps Required to Converge.}
\cref{fig:pretrain_learning_curve} depicts the number of training steps required to converge.

\begin{table*}[htb]
\centering
\begin{subtable}{0.95\linewidth}
\centering
\resizebox{\linewidth}{!}{
\begin{tabular}{p{15cm}}
\toprule
\textbf{A \pmb {($u_1$)}}: Did I do well on my test?
\textbf{B \pmb {($u_2$)}}: Do you want to know the honest answer?
\textbf{A \pmb {($u_3$)}}: Why wouldn't I want to know?
\textbf{B \pmb {($u_4$)}}: You had pretty bad scores.
\textbf{A \pmb {($u_5$)}}: Exactly what do you mean by bad?
\textbf{B \pmb {($u_6$)}}: You failed.
\textbf{A \pmb {($u_7$)}}: How'd I fail it?
\textbf{B \pmb {($u_8$)}}: There are a couple of reasons why you didn't pass.
\textbf{A \pmb {($u_9$)}}: What did I do wrong?
\textbf{B \pmb {($u_{10}$)}}: To sum it all up, you really just don't know how to drive.
\textbf{A \pmb {($u_{11}$)}}: Thanks. Will I be able to take a retest?
\textbf{B \pmb {($u_{12}$)}}: Sure you can , in about two and a half weeks. \\
\midrule
% \textbf{Target -} \pmb{$u_{10}$}; \textbf{Question:} What is or could be the \colorbox{lightgray}{\it emotional effect} of the target on the listener? \textbf{Inference:} The listener is disappointed with his score. \\
% \midrule
\textbf{Target -} \pmb{$u_{11}$}; \textbf{Question:} What is or could be the \colorbox{lightgray}{\it life goal} of the target? \textbf{Inference:} The speaker is hopeful of getting a re-test. \\
\midrule
\textbf{Target -} \pmb{$u_{12}$}; \textbf{Question:} What is or could be the \colorbox{lightgray}{\it physical requirement} of the target? \textbf{Inference:} The speaker has a good driving record. \\
\midrule
\textbf{Target -} \pmb{$u_{12}$}; \textbf{Question:} What is or could be the \colorbox{lightgray}{\it intention} of the target? \textbf{Inference:} The speaker is encouraging the listener to re-appear in the driving test. \\
\midrule  \midrule
\textbf{A \pmb {($u_1$)}}: David, do you like ice cream?
\textbf{B \pmb {($u_2$)}}: Yes I do, a lot!
\textbf{A \pmb {($u_3$)}}: Well, why don't we go get some today?
\textbf{B \pmb {($u_4$)}}: Sorry, I can not make it today as I have some other plans. \\
\midrule
\textbf{Target -} \pmb{$u_{4}$}; \textbf{Question:} What is the \colorbox{lightgray}{\it goal} of the speaker in the target? \textbf{Inference:} The speaker has to attend a meeting. \\
\midrule
\textbf{Target -} \pmb{$u_{4}$}; \textbf{Question:} What is the \colorbox{lightgray}{\it emotion} of the speaker in the target? \textbf{Inference:} The speaker is disappointed as he is unable to go for ice cream. \\
\midrule  \midrule
\textbf{A \pmb {($u_1$)}}: David, do you like ice cream?
\textbf{B \pmb {($u_2$)}}: Yes I do, a lot!
\textbf{A \pmb {($u_3$)}}: Well, why don't we go get some today?
\textbf{B \pmb {($u_4$)}}: I can't wait. \\
\midrule
\textbf{Target -} \pmb{$u_{4}$}; \textbf{Question:} What is the \colorbox{lightgray}{\it goal} of the speaker in the target? \textbf{Inference:} The speaker and david are craving for ice cream. \\
\midrule
\textbf{Target -} \pmb{$u_{4}$}; \textbf{Question:} What is the \colorbox{lightgray}{\it emotion} of the speaker in the target? \textbf{Inference:} The speaker is excited to go to the ice cream shop. \\
\bottomrule
\end{tabular}
}
\end{subtable}
%\vspace{-1.1em}
\caption{Examples of zero-shot question types and inferences.}
\label{tab:zero-shot-transfer}
\end{table*}

\paragraph{Examples of Generated Outputs from the Pre-training Stage.}
We provide examples of inputs, ground truth and generated outputs by \model{} in \cref{tab:generations1}.

\paragraph{Examples of Generated Output for Multiview Contextual Commonsense inference.}
We provide a few examples where  \model{} makes the correct predictions while the baseline model makes commonsense mistakes in \cref{tab:generations_finetuned}. 
For example, in the first dialogue, the model needs to guess what will happen next after the speaker complains \textit{Aspirin is not strong enough}. The baseline model mistakenly selects option 4, suggesting \textit{the listener to visit the emergency room to get medicines}. Similarly, in the second example, the baseline model predicts that \textit{a thief pulled out a knife} will \textit{ask if he was okay} as the next movement. 
In the following example, the waiter is confirming if the guest wants to book the room which requires the room to be not occupied and again contradicts option 4 predicted by the baseline model. \model{} makes such commonsense mistakes much less compared to the baseline.  
The last example illustrates a case where \model{} makes a prediction that contains the words \textit{upstairs} but fails to understand the relative spatial information of the speaker and listener and as a result, makes a commonsense mistake. It suggests that the model's ability to do inference still needs to be improved. 

\paragraph{Zero-shot Transfer with \model{}}
We examine if \model{} is capable of performing zero-shot inferences on unseen questions beyond the pre-training corpus. We show some examples of such inferences in \cref{tab:zero-shot-transfer}. \model{} provides correct inferences for \textit{life goal, physical requirement}, and \textit{intention} dimension in the first dialogue for different target utterances. The second and third dialogue contexts are constructed in a way such that the first three utterances are identical and the fourth utterance is different. We then ask questions about \textit{goal} and \textit{emotion} of the speaker for the fourth utterance. \model{} again generates accurate inferences for the questions. The inferences also change appropriately based on the distinct fourth utterances in the two dialogues.

\begin{table*}[!h]
\small
\centering
\resizebox{\linewidth}{!}{
\begin{tabular}{llL{8.25cm}L{3.5cm}L{3.5cm}}
\toprule
\textbf{Group} & \textbf{\# } & \textbf{Input} & \textbf{Reference} & \textbf{Generated Output}\\
\midrule

\multirow{10}{*}{PO} & (i) & What is or could be the cause of target? <sep> target: Drive slowly, David. You could have an accident. <sep> context: $x$ & David is driving very fast to flaunt his driving skills to the speaker. & The speaker is warning david not to drive too fast. \\

\cmidrule{2-5} & (ii) & For which utterance in the context the cause is the following: David is driving very fast to flaunt his driving skills to the speaker. <sep> context: $x$ & Drive slowly, David. You could have an accident. & You can count on me. I have been driving for years. \\

\cmidrule{2-5} & (iii) & target: Drive slowly, David. You could have an accident. <sep> The cause of the target: David is driving very fast to flaunt his driving skills to the speaker. <sep> What is the subsequent event of the target? <sep> context: $x$ & David ignores the speaker’s advice and continues driving with the same pace. & The speaker warns david that if he drives too fast he will get into an accident. \\

\midrule 

\multirow{17}{*}{SCAO} & (i) & What is or could be the cause of target? <sep> target: Drive slowly, David. You could have an accident. <sep> (0) David drives very slowly to flaunt his walking skills to the speaker. (1) David drives very slowly to flaunt his driving skills to the speaker. (2) David is driving very slowly to flaunt his driving skills to the speaker. (3) David is driving very fast to flaunt his driving skills to the speaker. (4) David walks very fast to flaunt his driving skills to the speaker. <sep> context: $x$ & David is driving very fast to flaunt his driving skills to the speaker. & David is driving very fast to flaunt his driving skills to the speaker. \\

\cmidrule{2-5} & (ii) & answer: David is driving very fast to flaunt his driving skills to the speaker. <sep> target: Drive slowly, David. You could have an accident. <sep> context: $x$ & cause & subsequent event \\

\cmidrule{2-5} & (iii) & The cause of the target: David is driving very fast to flaunt his driving skills to the speaker. <sep> target options: Drive slowly, David. You could have an accident. <utt> Look out! Red light! <utt> It doesn't matter. It is late. There is no one around. <utt> You can count on me. I have been driving for years. <sep> context: $x$ & Drive slowly, David. You could have an accident. & You can count on me. I have been driving for years. \\

\midrule 

\multirow{12}{*}{CO} & (i) & For which utterance in the context the cause is related to the following concepts: drive, flaunt, driving, skill, speaker <sep> context: $x$ & Drive slowly, David. You could have an accident. & You can count on me. I have been driving for years. \\

\cmidrule{2-5} & (ii) & For which utterance in the context the cause is the following: David is driving very fast to flaunt his driving skills to the speaker. <sep> concept: drive, accident <sep> context: $x$ & Drive slowly, David. You could have an accident. & Drive slowly, David. You could have an accident. \\

\cmidrule{2-5} & (iii) & What is or could be the cause of target? <sep> concepts in the target: drive, accident <sep> context: $x$ & David is driving very fast to flaunt his driving skills to the speaker. & David was driving at a high speed. \\

\cmidrule{2-5} & (iv) & What is or could be the cause of target? <sep> target: Drive slowly, David. You could have an accident. <sep> concepts in the answer: drive, flaunt, driving, skill, speaker <sep> context: $x$ & David is driving very fast to flaunt his driving skills to the speaker. & David was driving fast and flaunting his driving skills to the speaker. \\

\midrule 

\multirow{6}{*}{DO} & (i) & target: Drive slowly, David. You could have an accident. <sep> corrupted concepts: drive, driving, flaunt, speaker <sep> context: $x$ <sep> concepts in the answer:  & drive, flaunt, driving, skill, speaker & speaker, flaunt, driving, skill, drive \\

\cmidrule{2-5} & (ii) & answer: David is driving very fast to flaunt his driving skills to the speaker. <sep> corrupted concepts: drive <sep> context: $x$ <sep> concepts in the target:  & drive, accident & drive, accident \\

\midrule 

\multirow{13}{*}{SO} & (i) & context: $x$ <sep> David is driving very fast to flaunt his driving skills to the speaker. <sep> A policeman caught david for breaking traffic rules. <sep> David was driving very fast and broked traffic rules. <sep> The speaker would tell the listener to apply brakes. <sep> David ignores the speaker’s advice and continues driving with the same pace. <sep> David is confident in his driving skills. <sep> The speaker is driving with overconfidence that leads him to miss the traffic signal. & 0 6 5 4 1 2 3 & 6 0 1 3 5 4 2 \\

\cmidrule{2-5} & (ii) & B: You can count on me. I have been driving for years. <utt> A: Look out! Red light! <utt> B: It doesn't matter. It is late. There is no one around. <utt> A: Don't let the police catch you. Oh, David, that's a policeman. He is waving over us. <utt> A: Drive slowly, David. You could have an accident. & 4 0 1 2 3 & 4 0 1 2 3 \\
\bottomrule
\end{tabular}
}
\caption{An example of input, reference, and generated output triplets for the various groups of objective functions (\Cref{sec:pretrain-objective-groups}) from a dialogue $\mathcal{D}$.
PO, SCAO, CO, DO, and SO refers to the Primary Objectives, Single Correct Answer Objectives, Concept-Based Objectives, Denoising Objectives, and Sorting Based Objectives, respectively.
The outputs are generated from the pretrained DIALECT model. The context placeholder $x$ is the concatenation of the utterances in the dialogue $\mathcal{D}$, which is the following string: \textrm{A: Drive slowly, David. You could have an accident. <utt> B: You can count on me. I have been driving for years. <utt> A: Look out! Red light! <utt> B: It doesn't matter. It is late. There is no one around. <utt> A: Don't let the police catch you. Oh, David, that's a policeman. He is waving over us.}}
\label{tab:generations1}
\end{table*}
\begin{table*}[!hp]
%\small
\centering
\resizebox{\linewidth}{!}{
\begin{tabular}{L{7.5cm}L{6.5cm}C{1.5cm}C{1.5cm}C{1.6cm}}
\toprule
 \textbf{Context}  &  \textbf{Relation + Answers} &\textbf{Label} & \textbf{\model{}} & \textbf{T5-Large}\\
\midrule
A: Wake up. It's almost eight o'clock.
B: No, please. Let me sleep on! I couldn't get to sleep until 3 o'clock this morning.
A: Why? What's wrong with you?
B: I felt pain all over my body. Can you get me some medicine?
A: Will aspirin do?
\textbf{B: No, aspirin isn't strong enough.}
A: Then I can do nothing but call for a doctor.
& \textbf{Subseq:} (0) The speaker would tell the listener to visit the doctor to get some better medicines. (1) The speaker would tell the listener to call the doctor who would prescribe them medicine. (2) The speaker would tell the listener to call the doctor to see if they could get some more medicine. (3) The speaker would tell the listener to visit the medical store nearby to get some better medicines. (4) The speaker would tell the listener to visit the emergency room to get some better medicines.
& 0, 1, 3 
& 0, 1, 3
& 0, 3, 4 \\

\midrule

A: Hello, Joan. Why are you late today? You are never late for work.
B: No, I never. But ...
A: Wow! You coat's got very dirty! Did you fall?
\textbf{B: Yes, I had a terrible experience on the underground train. Listen to this! A man came up to me and pulled out a knife. He pointed it right at me!}
A: Oh, no! Are you all right? Did he hurt you?
B: No, he didn't hurt me, but he took my handbag.
A: Then what happened? What did you do?
B: I caught hold of his knife, and he pushed me to the floor.
A: Oh, no! Why did you catch hold of his knife? That's dangerous.
B: I don't know. I didn't think.
A: What did the other passengers do? Did they help you?
B: Yes, they did. Two men ran after the robber and held him.
A: Did the police come?
B: Yeah. The conductor called a policeman, and he took the robber to the police station.
A: Wow! What a story! Thank God you're all right.
& \textbf{Subseq:} (0) Joan would tell the listener that the thief asked him if he was okay. (1) Joan would tell the listener that the thief asked him to give him money and a watch. (2) Joan would tell the listener that the thief asked him to give him money and a cell phone. (3) Joan would tell the listener that the thief asked him to tell the police about the crime. (4) Joan told the listener that the thief asked him to hand over his keys. 
& 1, 2, 3
& 1, 2, 3 
& 0, 2 \\

\midrule

A: Hello , may I help you ?
B: Yes.We ' re interested in seeing the rooms for rent .
A: Oh , how nice.They ' re bright rooms and the house is very quiet .
B: A nice quiet house is exactly what we're looking for .
\textbf{A: Well , gentleman.Each room is \$ 40 a week if you think that's OK .}
B: That sounds just wonderful to us .
A: When do you want to move in ?
B: How about this afternoon ?
A: Fine . I'll be expecting you around two .
& \textbf{Prereq:} (0) The rooms showed to the person are currently unoccupied. (1) The rooms shown to the person are currently ready to be occupied. (2) The rooms they show are occupied. (3) The rooms showed to the person are full and occupied. (4) The rooms the person was looking in are currently occupied. 
& 0, 1 
& 0, 1
& 1, 4 \\

\midrule
A: Paul, is that you?  B: Yes, Mary. What can I do for you? A: Sorry to call you. But I just delivered my new computer. I am afraid I can't lift it by myself. Could you give me a hand to get it upstairs?  B: Sure. Could you just give me a minute to finish off what I am doing? \textbf{A: Yes, of course. But please hurry. The box is getting in the way.} B: Don't worry. I'll be right down.
& \textbf{Subseq:} (0) Paul will get down to pick up the computer from mary. (1) Paul will get downstairs to help mary in lifting the computer upstairs. (2) Paul will get upstairs to help mary in lifting the box upstairs. (3) Paul will get downstairs to help mary in lifting the box upstairs. (4) Mary will help paul lift the computer. 
& 1, 3
& 1, 2, 3
& 1, 3 \\

\bottomrule
\end{tabular}
}
\caption{Examples of the fine-tuning performance of \model{} and its comparison with T5-Large.}

\label{tab:generations_finetuned}
\end{table*}

% \newpage

\end{document}